\documentclass[11pt,a4paper]{article}
\usepackage[hyperref]{acl2018}
\usepackage{times}
\usepackage{latexsym}
\usepackage{url}
\usepackage{enumitem}
\usepackage{stfloats}
\usepackage{float}
\usepackage{mathtools}
\usepackage{graphicx}
\usepackage{epstopdf}
\usepackage{amsfonts}
\usepackage{amssymb}
\usepackage{CJK}
\usepackage{fixltx2e}
\usepackage{amsmath}
\usepackage{verbatim}
\usepackage{multirow}
\usepackage{subfigure}
\usepackage{pifont}% http://ctan.org/pkg/pifont
\usepackage{array}
\usepackage{fp}
\usepackage{makecell}
\usepackage{flushend}
\usepackage{cases}
\usepackage{makecell}
\usepackage{color}
\usepackage{bbm}
\usepackage{hyperref}
\newcommand{\thickhline}{\noalign{\hrule height 1pt}}
\aclfinalcopy

\makeatletter
\def\UrlAlphabet{%
	\do\a\do\b\do\c\do\d\do\e\do\f\do\g\do\h\do\i\do\j%
	\do\k\do\l\do\m\do\n\do\o\do\p\do\q\do\r\do\s\do\t%
	\do\u\do\v\do\w\do\x\do\y\do\z\do\A\do\B\do\C\do\D%
	\do\E\do\F\do\G\do\H\do\I\do\J\do\K\do\L\do\M\do\N%
	\do\O\do\P\do\Q\do\R\do\S\do\T\do\U\do\V\do\W\do\X%
	\do\Y\do\Z}
\def\UrlDigits{\do\1\do\2\do\3\do\4\do\5\do\6\do\7\do\8\do\9\do\0}
\g@addto@macro{\UrlBreaks}{\UrlOrds}
\g@addto@macro{\UrlBreaks}{\UrlAlphabet}
\g@addto@macro{\UrlBreaks}{\UrlDigits}
\makeatother

\title{Towards Explainable and Controllable Open Domain Dialogue Generation with Dialogue Acts}

\author{
	Can Xu$^\dag$, Wei Wu$^\dag$, Yu Wu$^\diamondsuit$~~~\\
$^ \dag$Microsoft Corporation, Beijing, China\\
$^\diamondsuit$State Key Lab of Software Development Environment, Beihang University, Beijing, China\\
\{wuwei,cax\}@microsoft.com, \{wuyu\}@buaa.edu.cn 
}

\begin{document}
	
	\begin{CJK*}{UTF8}{gbsn} 
		\maketitle
		
		\begin{abstract}
			We study open domain dialogue generation with dialogue acts designed to explain how people engage in social chat. To imitate human behavior, we propose managing the flow of human-machine interactions with the dialogue acts as policies. The policies and response generation are jointly learned from human-human conversations, and the former is further optimized with a reinforcement learning approach. With the dialogue acts, we achieve significant improvement over state-of-the-art methods on response quality for given contexts and dialogue length in both machine-machine simulation and human-machine conversation. %but also are capable of explaining why such achievements can be made. 
		\end{abstract}
		
		\section{Introduction}
		Recently, there is a surge of interest on dialogue generation for chatbots which aim to naturally and meaningfully converse with humans on open domain topics \cite{vinyals2015neural}. Although often called ``non-goal-oriented'' dialogue systems, such conversational agents are often built to keep users engaged in human-machine interactions as long as possible \cite{ram2018conversational}. While most of the existing effort is paid to generating relevant and diverse responses for static contexts  \cite{serban2015building,serban2017hierarchical,sordoni2015neural,li2015diversity}, it is not clear if relevance and diversity are sufficient to engagement in dynamic human-machine interactions, and if not, what else are needed to achieve the engagement.  

		In this work, we investigate the following problems: (1) how to understand human engagement in their social chat; (2) how to imitate such behavior in dialogue generation; (3) how to learn such a dialogue model;  and (4) if the model can control its responses in interactions and thus enhance user engagement.  
		
		We design dialogue acts that can describe how human behave regarding to conversational contexts in their social interactions. The dialogue acts, when applied to real data, give rise to an interesting finding that in addition to replying with relevance and diversity, people are used to driving their social chat by constantly switching to new contexts and properly asking questions. Such behavior is less explored before, and thus is difficult for the existing end-to-end learning methods to imitate. To mimic the behavior, we propose modeling open domain dialogue generation as an alternation of dialogue act selection and response generation where the dialogue acts control the types of the generated responses and thus manage the flow of interactions as policies. The model is learnt from large scale human-human dialogues tagged with a dialogue act classifier, and the policy of act selection is further optimized for long-term conversation through a reinforcement learning approach. Our model enjoys several advantages over the existing models: (1) the dialogue acts provide interpretation to response generation from a discourse perspective; (2) the dialogue acts enhance diversity of responses by expanding the search space from language to act $\times$ language; (3) the dialogue acts improve user engagement in human-machine interactions; and (4) the dialogue acts allow engineers to control their systems by picking responses from their desired acts. Evaluation results on large scale test data indicate that our model can significantly outperform state-of-the-art methods in terms of quality of generated responses regarding to given contexts and lead to long-term conversation in both machine-machine simulation and human-machine conversation.
		
		Our contributions in this work include:  (1) design of dialogue acts that represent human behavior regarding to conversational contexts and insights from analysis of human-human interactions; (2) joint modeling of  dialogue act selection and response generation in open domain dialogue generation; (3) proposal of learning the model through a supervised learning approach and a reinforcement learning approach; (4) empirical verification of the effectiveness of the model through automatic metrics, human annotations, machine-machine simulation, and human-machine conversation. 
		
		\begin{table*}[h]
			\small
			\centering
			%\vspace{-3mm}			
			\begin{tabular}{m{2cm}|m{6cm}|m{4.5cm}}
				\hline
				Dialogue Acts & Definitions& Examples\\	\hline

				Context Maintain Statement (CM.S)& A user or a bot aims to maintain the current conversational context (e.g., topic) by giving information, suggesting something, or commenting on the previous utterances, etc. &``\textbf{there are many good places in Tokyo.}'' after ''I plan to have a tour in Tokyo this summer.''.  \\ \hline
				Context Maintain Question (CM.Q) & A user or a bot asks a question in the current context. Questions cover 5W1H and yes-no with various functions such as context clarification, confirmation, knowledge acquisition, and rhetorical questions, etc.& ``\textbf{where are you going to stay in Tokyo?}'' after ``I plan to have a tour in Tokyo this summer.''.\\\hline
				Context Maintain Answer (CM.A) & A response or an answer to the previous utterances in the current context.& ``\textbf{this summer.}'' after ``when are you going to Tokyo?''.  \\ \hline
				Context Switch Statement (CS.S) & Similar to CM.S, but the user or the bot tries to switch to a new context (e.g., topic) by bringing in new content. & ``\textbf{I plan to study English this summer.}'' after ``I plan to have a tour in Tokyo this summer.''.\\ \hline
				Context Switch Question (CS.Q)& A user or a bot tries to change the context of conversation by asking a question. & ``\textbf{When will your summer vacation start?}'' after ``I plan to have a tour in Tokyo this summer.''\\ \hline
				Context Switch Answer (CS.A)& The utterance not only replies to the previous turn, but also starts a new topic. &``\textbf{I don't know because I have to get an A+ in my math exam.}'' after ``when are you going to Tokyo?''.\\ \hline
				Others (O)& greetings, thanks, and requests, etc.. &``\textbf{thanks for your help.}''\\ \hline
				
			\end{tabular}
			\caption{Definition of dialogue acts.}\label{table:def} 			
			%\vspace{-3mm}
		\end{table*}
		
		%\vspace{-1mm}
		\section{Dialogue Acts for  Social Engagement}\label{DiaAct}%\vspace{-3mm}
		%We first define dialogue acts, and then describe the data for learning and the insights we obtain from the data. Finally, we elaborate how we build the classifier with neural networks.  
		%\vspace{-1mm}
		\subsection{Definition of Dialogue Acts} \label{actdef}
		We define our dialogue acts by extending the $42$ tags \citep{jurafsky1997switchboard,stolcke2006dialogue} based on the DAMSL annotation scheme \citep{core1997coding}. Specifically, we merge some acts and define two high-level ones that describe how people behave regarding to conversational contexts in their interactions. As will be seen later, the extension brings us insights on engagement in social chat. Details of the dialogue acts are described in Table \ref{table:def}.

		The dialogue acts in Table \ref{table:def} are generally applicable to open domain dialogues from various sources in different languages such as Twitter, Reddit, Facebook, Weibo (\url{www.weibo.com}), and Baidu Tieba (\url{https://tieba.baidu.com/}), etc. 
		Existing annotated data sets (e.g., the Switchboard Corpus\footnote{\url{https://github.com/cgpotts/swda}}) do not have dialogue acts regarding to conversational contexts. Therefore, it is not clear how such dialogue acts depict human behavior in interactions, and there are no large scale data available for learning dialgoue generation with the dialogue acts either. To resolve these problems, we build a data set.  
		%\vspace{-3mm}
		\subsection{Data Set}\label{sec:dataset}
		We crawled $30$ million dyadic dialogues (conversations between two people) from Baidu Tieba. Baidu Tieba is the largest Reddit-like forum in China which allows users to communicate with each other through one posting a comment and the other one replying to the comment.  We randomly sample $9$ million dialogues as a training set, $90$ thousand dialogues as a validation set, and $1000$ dialogues as a test set. These data are used to learn a dialogue generation model later.  We employ the Standford Chinese word segmenter (\url{https://nlp.stanford.edu/software/tokenizer.shtml}) to tokenize utterances in the data. Table \ref{dataset} reports statistics of the data.
		\begin{table}[h]\small
			\centering
			%\vspace{-2mm}	
			
			\begin{tabular}{l|c|c|c}
				\thickhline
				&train&val&test\\ \hline
				$\#$ dialogues & 9M & 90k & 1000\\ \hline
				Min. $\#$  turns per dialogue  &3&5&5\\ \hline
				Max. $\# $  turns per dialogue   &50&50&50\\\hline
				Avg. $\#$ turns per dialogue &7.68&7.67&7.66\\\hline
				Avg. $\#$  words per utterance &15.81&15.89&15.74\\
				\thickhline
			\end{tabular}
		\caption{Statistics of the experimental data sets.}	\label{dataset}
			%\vspace{-5mm}
		\end{table}
		
		For dialogue act learning, we randomly sample $500$ dialogues from the training set and recruit $3$ native speakers to label dialogue acts for each utterance according to the definitions in Table \ref{table:def}. Table \ref{actlabelexample} shows a labeling example from one annotator. Each utterance receives $3$ labels, and the Fleiss' kappa of the labeling work is $0.45$, indicating moderate agreement among the labelers.    
		%\footnote{By default, the first utterance in a dialogue is labeled as CM.* except clear opening expressions such as ``hello'' or ``morning'', because the first utterance often follows the context of a post.} 
		
		\begin{table*}[h]
			\small
			%\vspace{-4mm}
			\centering	
			\begin{tabular}{l |l}
				\thickhline
				Turns & Dialogue Acts \\ \hline
				A: 万里长城很漂亮！ The Great Wall of China is beautiful! & CM.S\\ 
				B: 你在长城看日落了吗？ Did you see the sunset on the Great Wall? & CM.Q\\
				A: 是的，那是最漂亮的景色。 Yes, it's the most beautiful scenery.& CM.A\\
				B: 上次我去的时候人很多。 It was very crowded when I visited there last time & CS.S\\
				A: 我只待了一小会儿，人太多了！ I only stayed there for a while. Too many vistors! & CM.S\\
				\thickhline
			\end{tabular}
			\caption{An example of dialogue with labeled acts.}\label{actlabelexample}				
			%\vspace{-5mm}
		\end{table*}
		
		\subsection{Insights from the labeled data}\label{insight}
		The frequencies of the dialogue acts in terms of percentages of the total number of utterances in the labeled data are CM.S $55.8$\%, CM.Q $11.7$\%, CM.A $12.2$\%, CS.S $12.4$\%, CS.Q $4.8$\%, CS.A $2$\%, and O $1.1$\%. %The distribution indicates that the most common act people perform in chit-chat is making a statement to maintain the current conversational context, which is consistent with our intuition. 
		In addition to the numbers, we also get further insights from the data that are instructive to our dialogue generation learning:

		\textbf{Context switch} is a common skill to keep conversation going. In fact, we find that $78.2$\% dialogues contain at least one CS.* act. The average number of turns of dialogues that contain at least one CS.* is $8.4$, while the average number of turns of dialogues that do not contain a CS.* is $7$. When dialogues are shorter than $5$ turns, only $47$\% of them contain a CS.*, but when  dialogues exceed $10$ turns, more than $85$\% of them contain a CS.*. Because there are no specific goals in their conversations, people seldom stay long in one context. The average number of turns before context switch is $3.39$.  We also observed consecutive context switch in many dialogues ($43.7$\%). The numbers suggest dialogue generation with smooth context switch and moderate context maintenance.%, which is difficult for the existing methods to achieve. 	
		
		\textbf{Question} is an important building block in open domain conversation. In fact, $13.9$\% CM.* are CM.Q and the percentage is even higher in CS.* which is $20.27$\%. People need to ask questions in order to maintain contexts. The average number of turns of contexts with questions (i.e., consecutive CM.* with at least one CM.Q) is $3.92$, while the average number of turns of contexts without questions is only $2.95$. The observation indicates that a good dialogue model should be capable of asking questions properly, as suggested by \cite{li2016learning}. %Questions in contexts (i.e., CM.Q) are not necessarily followed by answers. In addition to CM.A ($64$\%), common actions after CM.Q include CM.S ($14$\%), CM.Q ($10$\%), and CS.S ($8$\%). 
		A further step to study human's questioning behavior is to look into types and functions of questions. We leave it as future work. 	
		%\item Questions in contexts (i.e., CM.Q) are not necessarily followed by answers. In addition to CM.A ($64$\%), common actions after CM.Q include CM.S ($14$\%), CM.Q ($10$\%), and CS.S ($8$\%). In open domain conversation, people sometimes just igore questions they are asked and continue what they have talked in previous turns or switch to a new topic, and sometimes ask another question to confirm some conditions or even express sarcasm (e.g., by rhetorical questions). The phenomenon demonstrates complexity of open domain conversation in real world.    	       

		The observations raise new challenges that are difficult for the existing end-to-end methods to tackle (e.g., smoothly interleaving context blocks with switch actions), and thus encourage us to create a new model. Before elaborating the model, we first  build a classifier that can automatically tag large scale dialogues with the dialogue acts. 
		
		\subsection{Dialogue Act Classification}\label{DiaActCla}
		We aim to learn a classifier $c$ from $\mathcal{D}_A = \{d_i\}^N_{i=1}$ where $d_i=\{(u_{i,1},a_{i,1}),\ldots,(u_{i, n_i}, a_{i,n_i})\}$ represents a dialogue with $u_{i,k}$ the $k$-th utterance and $a_{i,k}$ the labeled dialogue act. Given a new dialogue $d=\{u_1,\ldots, u_n\}$, $c$ can sequentially tag the utterances in $d$ with dialouge acts by taking $u_i$, $u_{i-1}$, and the predicted $a_{i-1}$ as inputs and outputting a vector $c(u_i, u_{i-1}, a_{i-1})$ where the $j$-th element representing the probability of $u_i$ being tagged as the $j$-th dialogue act.  
		
		We parameterize $c(\cdot, \cdot, \cdot)$ using neural networks. Specifically, $u_i$ and $u_{i-1}$ are first processed by bidirectional recurrent neural networks with gated recurrent units (biGRUs) \citep{chung2014empirical} respectively. Then the last hidden states of the two biGRUs are concatenated with an embedding of $a_{i-1}$ and fed to a multi-layer perceptron (MLP) to calculate a dialogue act distribution. Formally, suppose that $u_i=(w_{i,1}, \ldots, w_{i,n})$ where $w_{i,j}$ is the embedding of the $j$-th word, then the $j$-th hidden state of the biGRU is given by $h_{i,j}=[\overrightarrow{h}_{i,j};\overleftarrow{h}_{i,j}]$ where $\overrightarrow{h}_{i,j}$ is the $j$-th state of a forward GRU, $\overleftarrow{h}_{i,j}$ is the $j$-th state of a backward GRU, and $[\cdot;\cdot]$ is a concatenation operator. $\overrightarrow{h}_{i,j}$ and $\overleftarrow{h}_{i,j}$ are calculated by %\vspace{-1mm}		
		\begin{equation}\label{biGRU} \small
		\overrightarrow{h}_{i,j}=f_{\text{GRU}} (\overrightarrow{h}_{i,j-1}, w_{i,j}); \thickspace \overleftarrow{h}_{i,j}=f_{\text{GRU}} (\overleftarrow{h}_{i,j+1}, w_{i,j}).
		\end{equation} 
		Similarly, we have $h_{i-1,j}$ as the $j$-th hidden state of $u_{i-1}$. Let $e(a_{i-1})$ be the embedding of $a_{i-1}$, then $c(u_i,u_{i-1},a_{i-1})$ is defined by a two-layer MLP:%\vspace{-1.5mm}
		\begin{equation}\small
		c(u_i,u_{i-1},a_{i-1})=f_{\text{MLP}}([h_{i,n}; h_{i-1,n}; e(a_{i-1})]),%\vspace{-1.5mm}
		\end{equation}
		where we pad zeros for $u_0$ and $a_0$ in $c(u_1,u_0,a_0)$. We learn $c(\cdot, \cdot, \cdot)$ by minimizing cross entropy with $\mathcal{D}_A$. Let $p_j(a_i)$ be the probability of $a_i$ being the $j$-th dialogue act and $c(u_i,u_{i-1},a_{i-1})[j]$ be the $j$-th element of $c(u_i,u_{i-1},a_{i-1})$, then the objective function of learning is formulated as %\vspace{-1.5mm}
		\begin{equation}\label{obj4actlearn} \small
		-\sum_{i=1}^N \sum_{k=1}^{n_i} \sum_{j=1}^7 p_j(a_{i,k}) \log(c(u_{i,k},u_{i,k-1},a_{i,k-1})[j]).%\vspace{-1.5mm}
		\end{equation}%\vspace{-3mm}
		
		We randomly split the labeled dialogues as $400$/$30$/$70$ dialogues with $3280$/$210$/$586$ utterances for training/validation/test. Details of model training are given in Appendix. The learned classifier achieves an accuracy of $70.1$\% on the test data. We employ it to tag the training, validation, and test sets in Table \ref{dataset}.

		\begin{figure*}[t]	
			\centering
			\subfigure[generation network]{	\includegraphics[width=5.5cm,height=3.5cm]{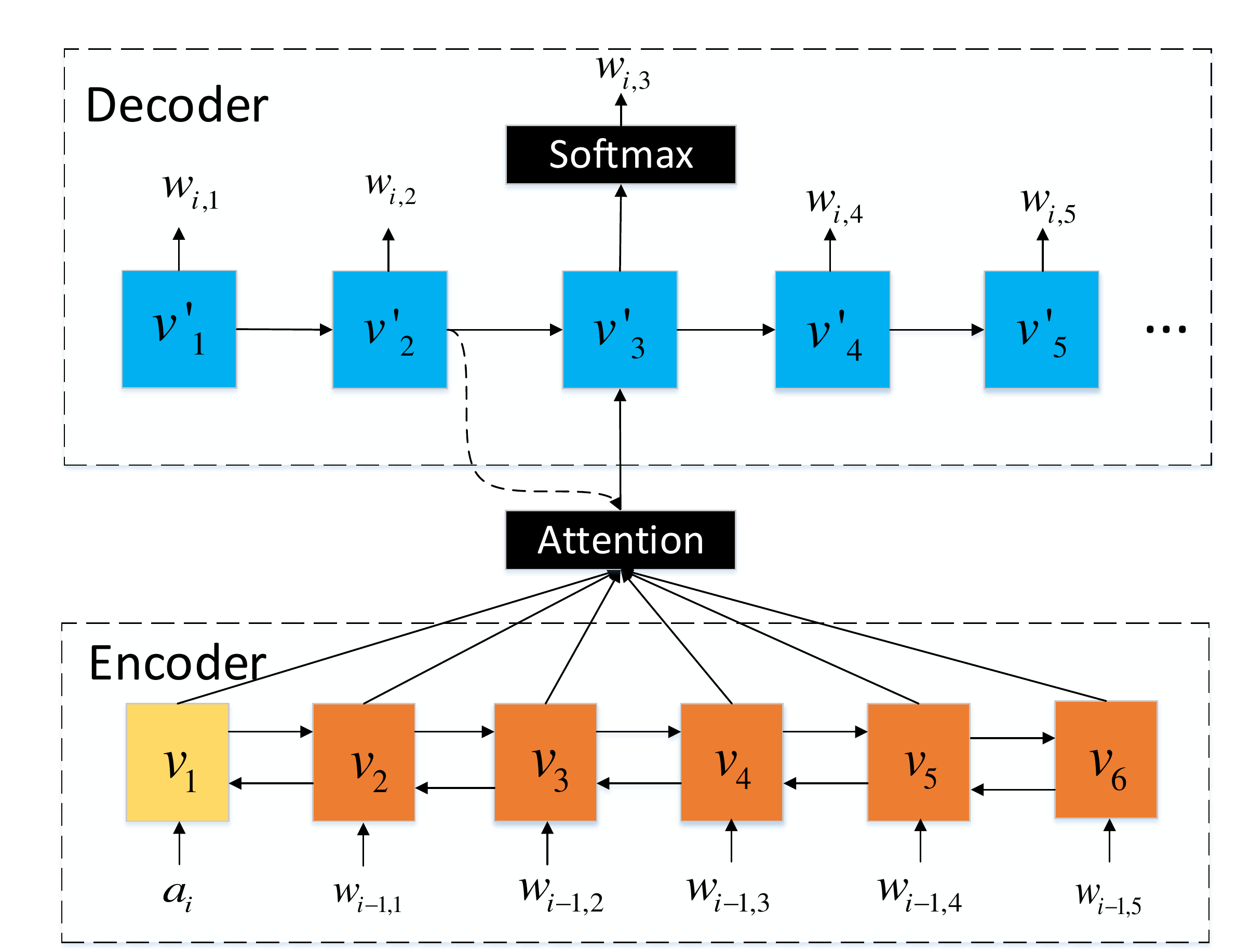}}	
			\subfigure[policy network]
			{
				\includegraphics[width=7cm,height=3.5cm]{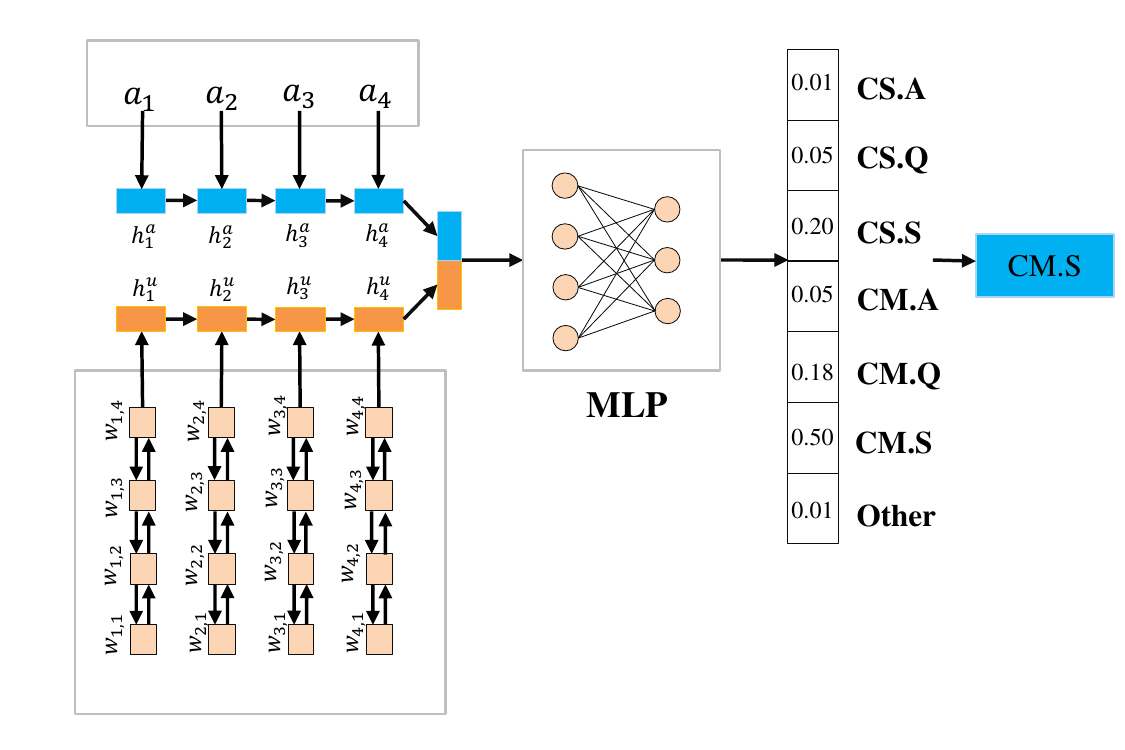}
			}
						%\vspace{-3mm}
			\caption{Policy network and generation network.} \label{fig:arc}	
			%\vspace{-4mm}
		\end{figure*}

		%\vspace{-3mm}
		\section{Dialogue Generation Model}
		\label{DiaMod}
		%We present dialogue generation learning using large scale dialogues tagged with dialogue acts. Then, we describe model optimization with reinforcement learning for long-term conversation.  
		
		\subsection{Supervised Learning}
		\label{SL}
		We aim to learn a dialogue generation model $g$ from  $\mathcal{D} = \{d_i\}^N_{i=1}$ where $d_i=\{(u_{i,1},a_{i,1}),\ldots,(u_{i, n_i}, a_{i,n_i})\}$ refers to a human-human dialogue with $u_{i,k}$ the $k$-th utterance and $a_{i,k}$ the dialogue act tagged by the classifier in Section \ref{DiaActCla}. Given $s_{i}=\{(u_1,a_1),\ldots,(u_{i-1}, a_{i-1})\}$ as a new dialogue session, $g(s_{i})$ can generate a response as the next turn of the dialogue.
		
		Our dialogue model consists of a policy network and a generation network.  A dialogue act is first selected from the policy network according to the conversation history, and then a response is generated from the generation network based on the conversation history and the dialogue act. Formally, the dialogue model can be formulated as %\vspace{-1.5mm}
		\begin{equation}\label{genmodel} \small
		g(s_{i})=p_r(r_i | s_{i}, a^\star_i), %\vspace{-3mm}
		\end{equation}
		where $a^\star_i=\arg\max_{a_i\in \mathbb{A}} \thickspace p_a(a_i |s_{i})$ is the selected dialogue act for the $i$-th turn, and $r_i$ is the response.  $p_a$ and $p_r$ are the policy network and the generation network respectively. $\mathbb{A}$ is the space of dialogue acts. 
		
		Figure \ref{fig:arc}(b) shows the architecture of the policy network. The utterance sequence and the act sequence are encoded with a hierarchical encoder and a GRU encoder respectively. Then, the last hidden states of the two encoders are concatenated and fed to an MLP
		to calculate a probability distribution of dialogue acts for the next turn. Formally, $\forall u_j \in s_{i}$, $u_j$ is first transformed to hidden vectors $\{h^u_{j,k}\}_{k=1}^{n'_j}$ through a biGRU parameterized as Equation (\ref{biGRU}). Then, $\{h^u_{j,n'_j}\}_{j=1}^{i-1}$ is processed by a GRU parameterized as $t_k=f^u_{\text{GRU}}(t_{k-1}, h^u_{k, n'_k})$. In parallel, $\{a_1,\ldots, a_{i-1}\}$ is transformed to $\{h^a_k\}_{k=1}^{i-1}$ by $h^a_{k}=f^a_{\text{GRU}} (h^a_{k-1}, e(a_{k}))$. $p_a(a_i | s_{i})$ is then defined by%\vspace{-1mm}
		\begin{equation} \small
		p_a(a_i | s_{i})= f_{\text{MLP}}([t_{i-1};h^a_{i-1}]). %\vspace{-1mm}
		\end{equation}  

		We build the generation network in a sequence-to-sequence framework. Here, we simplify $p_r(r_i | s_{i}, a_i)$ as $p_r(r_i| a_i, u_{i-1}, u_{i-2})$ since decoding natural language responses from long conversation history is challenging. Figure \ref{fig:arc}(a) illustrates the architecture of the generation network. The only difference from the standard encoder-decoder architecture with an attention mechanism \cite{bahdanau2014neural} is that in encoding, we concatenate $u_{i-1}$ and $u_{i-2}$, and attach $a_i$ to the top of the long sentence as a special word. The technique here is similar to that in zero-shot machine translation \citep{johnson2016google}. Formulation details are given in Appendix.
		
		The dialogue model is then learned by minimizing the negative log likelihood of $\mathcal{D}$: %\vspace{-2mm}
		\begin{equation} \small
		-\sum_{i=1}^N \sum_{k=1}^{n_i} [\log(p_r(u_{i,k} | d_{i,<k}, a_{i,k}))+\log(p_a(a_{i,k} | d_{i,<k}))],
		\end{equation}
		where $d_{i,<k}=\{(u_{i,1},a_{i,1}),\ldots,(u_{i, k-1}, a_{i,k-1})\}$. Through supervised learning, we fit the dialogue model to human-human interactions in order to learn their conversational patterns and human language. However, supervised learning does not explicitly encourage long-term conversation (e.g., $45.35$\% dialogues in our training set are no more than $5$ turns), and the policy network is optimized without awareness of what is going to happen in the future when a dialogue act is selected. This motivates us to further optimize the model through a reinforcement learning approach.      
		%\vspace{-3mm}
		\subsection{Reinforcement Learning}\label{RL}
		We optimize the dialogue model through self-play \citep{li2016deep,lewis2017deal} where we let two models learned with the supervised approach talk to each other in order to improve their performance. In the simulation, a dialogue is initialized with a message sampled from the training set. Then, the two models continue the dialogue by alternately taking the conversation history as an input and generating a response (top one in beam search) until $T$ turns ($T=20$ in our experiments).  
		
		To speed up training and avoid generated responses diverging from human language, we fix the generation network and only optimize the policy network by reinforcement learning. Thus, the policy in learning is naturally defined by the policy network $p_a(a_i| s_{i})$ with $s_{i}=\{(u_1,a_1),\ldots,(u_{i-1}, a_{i-1})\}$ a state and $a_i$ an action. We define a reward function $r(a_i, s_{i})$ as%\vspace{-1.5mm}
		\begin{equation}\label{reward} \small
		r(a_i, s_{i})=\alpha \mathbb{E}[len(a_i, s_{i})] +\beta \mathbb{E} [rel(a_i, s_{i})],%\vspace{-1.5mm}
		\end{equation}
		where $\mathbb{E}[len(a_i, s_{i})]$ is the expected dialogue length after taking $a_i$ under $s_{i}$,  $\mathbb{E} [rel(a_i, s_{i})]$ is the expected response relevance within the conversation, $\alpha=0.67$, and $\beta=0.33$. Through Equation (\ref{reward}), we try to encourage actions that can lead to long (measured by $\mathbb{E}[len(a_i, s_{i})]$) and reasonable (measured by $\mathbb{E} [rel(a_i, s_{i})]$) conversations.

		To estimate $\mathbb{E}[len(a_i,s_i)]$ and $\mathbb{E}[rel(a_i,s_i)]$, we fix $(s_i, a_i)$ and construct a dialogue set $\{d'_{i,j}\}_{j=1}^N$ ($N=10$ in our experiments) by sampling after $(s_i, a_i)$ with self-play. $\forall j$, $d'_{i,j}=(s_i, u_{j, i+1},\ldots, u_{j, n_{i,j}})$ where $\forall k$, $u_{j,i+k}$ is randomly sampled from the top 5 beam search results of $p_r$ according to Equation (\ref{genmodel}). Inspired by \citep{li2016deep}, we terminate a simulated dialogue if (1) $cosine(e(u_{i-1}),e(u_i)) > 0.9$ \&\& $cosine(e(u_i)),e(u_{i+1})) > 0.9$, or (2) $cosine(e(u_{i-1}),e(u_{i+1})) > 0.9$, or (3) the length of the dialogue reaches $T$, where $e(\cdot)$ denotes the representation of an utterance given by the encoder of $p_r$. Condition (1) means three consecutive turns are (semantically) repetitive, and Condition (2) means one agent gives repetitive responses in two consecutive turns. Both conditions indicate a high probability that the conversation falls into a bad infinite loop. $\mathbb{E}[len(a_i, s_{i})]$ and $\mathbb{E} [rel(a_i, s_{i})]$ are then estimated by%\vspace{-3mm}
		\begin{align*} 
		\small
		& \mathbb{E}[len(a_i, s_{i})]=\frac{1}{N} \sum_{j=1}^N n_{i,j} ; \\ &\mathbb{E}[rel(a_i, s_{i})]=\frac{1}{N} \sum_{j=1}^N \frac{1}{n_{i,j}} \sum_{k=1}^{n_{i,j}} m(d_{i,j<k}, u_{j,k}),%\vspace{-3mm}	
		\end{align*} 
		where $d_{i,j<k}=(u_1, \ldots, u_{j,k-1})$, and $m(\cdot,\cdot)$ is the dual LSTM model proposed in \citep{lowe2015ubuntu} which measures the relevance between a response and a context. We train $m(\cdot,\cdot)$ with the $30$ million crawled data through negative sampling. 
		The objective of learning is to maximize the expected future reward:%\vspace{-3mm}
		\begin{equation} \small
		\mathcal{J}(\theta)=\mathbb{E}[\sum_{i=1}^T r(a_i, s_i)].%\vspace{-3mm}
		\end{equation}
		The gradient of the objective is calculated by Reinforce algorithm \citep{williams1992simple}: %\vspace{-2mm}
		\begin{equation} \small
		\partial_\theta{\mathcal{J}} \approx \sum_{t=1}^T \partial_\theta{\log(p_a(a_t |s_t))} \big(\sum_{i=t}^T (r(a_i,s_i)-b_t)\big), %\vspace{-2mm}
		\end{equation}
		where the baseline $b_t$ is empirically set as $\frac{1}{|\mathbb{A}|} \sum_{a_t \in \mathbb{A}} r(a_t,s_t)$. %with $\mathbb{A}$ the space of actions.  
		%\vspace{-3mm}
		\section{Experiment}%\vspace{-3mm}
		\subsection{Experiment Setup}
		Our experiments are conducted with the data in Table \ref{dataset}. The following methods are employed as baselines: (1) \textbf{S2SA}: sequence-to-sequence with attention \citep{bahdanau2014neural} in which utterances in contexts are concatenated as a long sentence. We use the implementation with Blocks (\url{https://github.com/mila-udem/blocks}); (2) \textbf{HRED}: the hierarchical encoder-decoder model in \citep{serban2015building} implemented with the source code available at (\url{https://github.com/julianser/hed-dlg-truncated}); (3) \textbf{VHRED}: the hierarchical latent variable encoder-decoder model in \citep{serban2017hierarchical} implemented with the source code available at (\url{https://github.com/julianser/hed-dlg-truncated}); and (4) \textbf{RL-S2S}: dialogue generation with reinforcement learning \citep{li2016deep}. We implement the algorihtm by finishing the code at (\url{https://github.com/liuyuemaicha/Deep-Reinforcement-Learning-for-Dialogue-Generation-in-tensorflow}).   
		
		All baselines are implemented with the recommended configurations in the literatures. We denote our Dialogue Act aware Generation Model with only Supervised Learning as SL-DAGM, and the full model (supervised learning + reinforcement learning) as RL-DAGM. Implementation details are given in Appendix. %including the parameter settings and the dull responses in RL-S2S, are given in the supplementary material. 
		%\vspace{-3mm}
		\subsection{Response Generation for Given Contexts}
		The first experiment is to check if the proposed models can generate high-quality responses regarding to given contexts. To this end, we take the last turn of each test dialogue as ground truth, and feed the previous turns as a context to different models for response generation. Top one responses from beam search (beam size$=20$) of different models are collected, randomly shuffled, and presented to $3$ native speakers to judge their quality. Each response is rated by the three annotators under the following criteria: \textbf{2}: the response is not only relevant and natural, but also informative and interesting; \textbf{1}: the response can be used as a reply, but might not be informative enough (e.g.,``Yes, I see'' etc.); \textbf{0}: the response makes no sense, is irrelevant, or is grammatically broken.   
		
		\begin{table}[h]
			\small
			\centering %\vspace{-2mm}
			%\scalebox{0.8}{
				\subtable[Human annotations. Ratios are calculated by combining labels from the three judges. 	\label{exp:human}]	 {	\begin{tabular}{l|c|c|c|c}
						\hline
						& 0 & 1 & 2 & Kappa \\ \hline
						
						S2SA & 0.478 & 0.478 & 0.044 &0.528 \\ 
						HRED  & 0.447 & 0.456 & 0.097 & 0.492\\ 
						VHRED &  0.349 & 0.471 & 0.180 & 0.494\\ 	 		 		
						RL-S2S & 0.393 & 0.462 & 0.142 &0.501 \\ \hline
						SL-DAGM &0.279&0.475&0.244 & 0.508\\ 
						RL-DAGM &0.341&0.386&0.273 &0.485  \\ \hline
						
						\hline
					\end{tabular}
					
				}			
			%}
			%\scalebox{0.8}{
				\subtable[Average dialogue length in machine-machine and human-machine conversations. 	\label{exp:simul}]{	\begin{tabular}{l|c|c}
						\hline
						& Machine-Machine & Human-Machine \\ \hline

						RL-S2S & 4.36 & 4.54  \\ \hline
						SL-DAGM & 7.36 & 5.24\\ 
						RL-DAGM & 7.87 & 5.58 \\ \hline
						
						\hline
					\end{tabular}	
					
				}		
			%}          
			     %\vspace{-2mm}
			\caption{Evaluation Results}\label{evalres}
			%\vspace{-3mm}
		\end{table}

		Table \ref{evalres} (a) summarizes the annotation results. Improvements from our models over the baseline methods are statistically significant (t-test, p-value $<0.01$). Besides human annotations, we also compare different models using automatic metrics with the ground truth. These metrics include BLEU \citep{papineni2002bleu}, %which measures term overlap of two responses; 
		embedding based metrics \citep{liu2016not} such as Embedding Average (Average), Embedding Extrema (Extrema), and Embedding Greedy (Greedy), %which measure similarity of two responses in a semantic space; 
		and ratios of distinct unigrams (distinct-1) and bigrams (distinct-2) in the generated responses which are employed in \citep{li2015diversity} to measure response diversity.  Table \ref{autonum} reports the results.
		
		\begin{table*}[h]
			\small
			\centering %\vspace{-1mm}	
			
			\begin{tabular}{l|c|c|c|c|c|c|c}
				\hline
				& BLEU-1 & BLEU-2 &   Average&  Extrema &  Greedy &  Distinct-1 &  Distinct-2\\ \hline
				
				S2SA & 4.67 & 1.18  & 21.45&16.68& 21.53& 0.033 & 0.069\\ 
				HRED  & 3.70 & 1.06  & 16.87& 13.58& 20.15 & 0.062 & 0.139\\ 
				VHRED &  6.10 & 1.76 &20.83&16.17&21.36&0.079 & 0.225\\ 	 		 		
				RL-S2S &  5.57 & 1.83  &20.72&16.73&20.64 & 0.100 & 0.213\\ \hline
				SL-DAGM & 6.23 & 2.07 &20.68&16.42&21.52 & \textbf{0.200} &\textbf{0.466}\\ 
				RL-DAGM &6.77&2.12&21.18&16.97&21.76& \textbf{0.223} & \textbf{0.503}\\ \hline
				
				\hline
				
			\end{tabular}
			%\vspace{-1mm}
			\caption{Automatic evaluation results. Numbers in bold mean that improvement from the model on that metric is statistically significant over the baseline methods (t-test, p-value $<0.01$). }		\label{autonum} 			
		%\vspace{-2mm}
		\end{table*}

		We can see that diversity of responses is significantly improved with the dialogue acts. This is supported by the much more $2$ responses from the two models in Table \ref{evalres} (a) and the significant improvement on distinct n-grams in Table \ref{autonum}. The reason is that we search a response not only from a language space, but also from an act space. The dimension of dialogue acts provides further variations to the generated responses. On the other hand, due to the diversity, responses from our models may diverge from the ground truth sometimes. This is why improvements on other automatic metrics are not significant. To further explain the advantages of our models, we show an example in Table \ref{case}. Besides responses from the dialogue acts selected by our models, we also show responses from other reasonable but not selected acts. With the dialogue acts, the generated responses become really rich, from confirmation (CM.Q) to an open question (CS.Q) and then to a long informative statement (CS.S). More importantly, the dialogue acts let us know why we have such responses: both SL-DAGM and RL-DAGM try to switch to new topics (e.g., Xiamen, noodle, and plan etc.) in order to continue the conversation. One can also change the flow of the conversation by picking responses from other dialogue acts. The example demonstrates that besides good performance, our models enjoy good interpretability and controllability as well.  We show more such examples in Appendix.
		
		To further understand how the dialogue acts affect response generation, we collect generated responses from a specific dialogue act for the contexts of the test dialogues, and characterize the responses with the following metrics: (1) distinct-1 and distinct-2; (2) words out of context (OOC): ratio of words that are in the generated responses but not contained by the contexts; and (3) average length of the generated responses (Ave Len). 
				
		Table \ref{RDA} reports the results. %\footnote{We omit the dialogue act O, as only $1.1$\% of the labeled data are O and it is difficult for neural networks to capture the characteristics of text from such a few data. In practice, one can generate text for O by editorial.}. 
		In general, responses generated from CS.* are longer, more informative, and contain more new words than responses generated from CM.*, which has been illustrated in Table \ref{case}. Another interesting finding is that statements and answers are generally more informative than questions in both CS.* and CM.*. In addition to these metrics, we also calculate BLEU scores and embedding based metrics, but do not observe significant difference among responses from different dialogue acts. The reason might be that these metrics are based on comparsion of the generated responses and human responses, but human responses in the test set are inherently mixture of responses from different dialogue acts.

		\begin{table*}[t]
			\small
			\footnotesize
			\centering 

			\begin{tabular}{p{0.35\linewidth}|m{0.6\linewidth}}
				\hline
				Context &  Responses \\	\hline
				
				\begin{tabular}{b{4.5cm}}
					一起吃晚餐?  $\Rightarrow$ 中饭好吗？ $\Rightarrow$  中饭只能在公司吃 。$\Rightarrow$  那我不能来了。我在休假。 \\
					\\
					Have dinner together?   $\Rightarrow$  how about lunch?  $\Rightarrow$  I can only have my lunch at company.  $\Rightarrow$  Then I cannot join you because I am in my vacation.   
				\end{tabular}
				&
				\begin{tabular}{m{8cm}}
					
					\textbf{S2SA}: 我也是这么想的。 I think so \\
					\textbf{HRED}： 放假了啊？ You are already in vacation? \\
					\textbf{VHRED}： 哈哈哈。 Haha. \\
					\textbf{RL-S2S}： 我已经在吃了。 I am having lunch now. \\
					\textbf{SL-DAGM} ：好吧，我刚从厦门回来，想在食堂吃碗面。 OK. I am just back from Xiamen, and want to have noodle in cafeteria. (\textbf{CS.S})\\ 
					\textbf{RL-DAGM} ：放假有什么安排。 What are you going to do for your vacation? (\textbf{CS.Q})\\   
					\textbf{CM.Q}：放假了吗？ You are already in vacation? \\
					\textbf{CM.S}：我还以为你没休假呢。 I thought you were at work.\\ 
				\end{tabular}\\
				\hline
			\end{tabular}
			%\vspace{-3mm}
			\caption{An example of response generation. Utterances in the context are split by ``$\Rightarrow$''. }		\label{case} 			
			%\vspace{-5mm}	
		\end{table*} 
	
	\begin{table*}[h]
	\small
	\centering
	
	\begin{tabular}{l|c|c|c|c}
		\hline
		& Distinct-1 &  Distinct-2 & OOC & Ave Len\\ \hline
		
		CM.S & 0.114 &  0.262 & 0.091& 5.57\\ 
		CM.Q  & 0.092 & 0.220  & 0.038& 5.21\\ 
		CM.A &  0.119 & 0.269 &0.094&5.58\\ 	 		 		
		CS.S & 0.250 & 0.521  &0.168&8.21\\
		CS.Q & 0.223 & 0.460 & 0.152& 5.85\\ 
		CS.A &0.244& 0.500& 0.166&8.42\\ \hline
		
		\hline
	\end{tabular}
	%\vspace{-1mm}
	\caption{Characteristics of the generated responses from different dialogue acts. }	\label{RDA} 
	%\vspace{-3mm}	
\end{table*}	
	
		%\vspace{-3mm}
		\subsection{Engagement Test}
		
	\begin{figure*}[t]
	\small		
	\centering
	{
		\includegraphics[width=9.8cm,height=3.5cm]{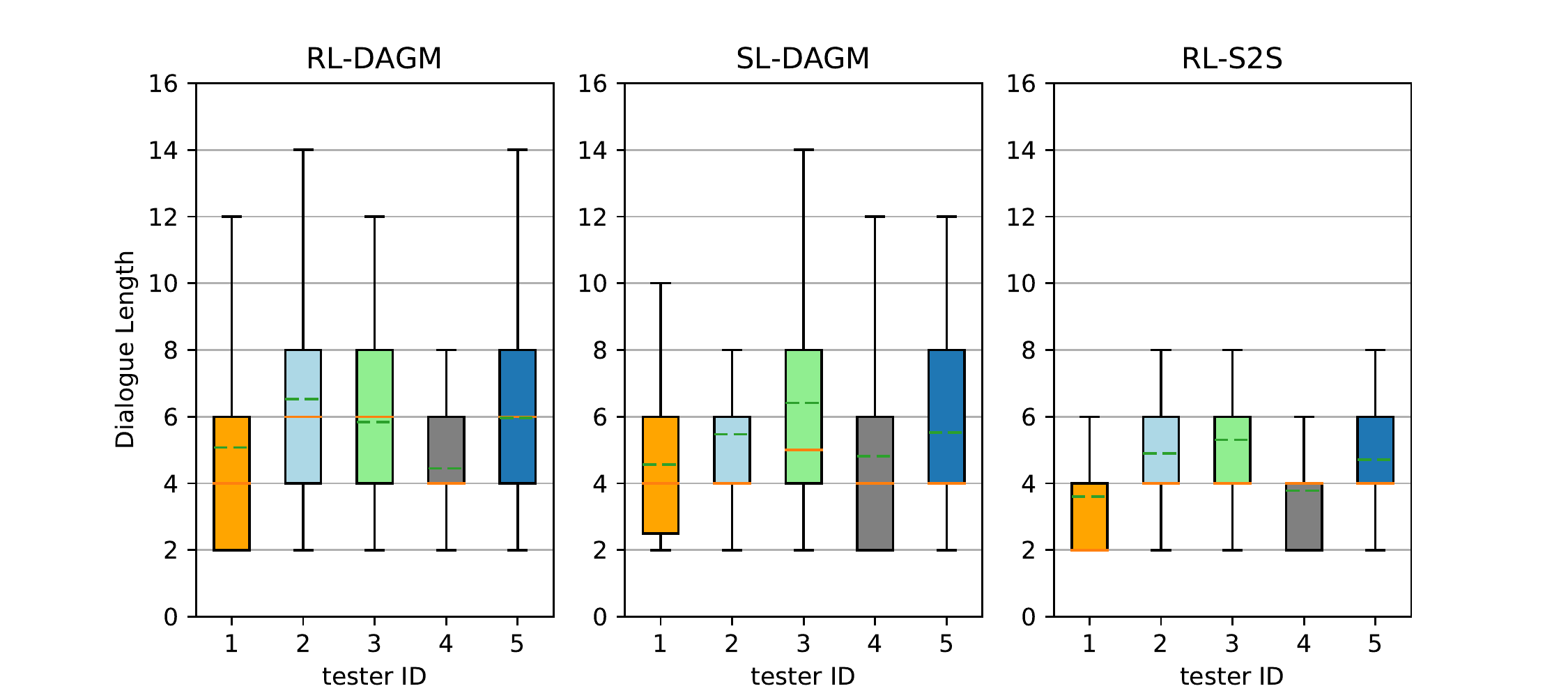}
	}
    %\vspace{-3mm}
	\caption{Average dialogue length of human-machine conversation in terms of different testers.} \label{boxgraph} %\vspace{-2mm}
\end{figure*}	
			
		Secondly, we study conversation engagement with the proposed models. Experiments are conducted through machine-machine simulation and human-machine conversation. In both experiments, we compare SL-DAGM and RL-DAGM with RL-S2S, as RL-S2S is the only baseline optimized for future success. Responses from all models are randomly sampled from the top $5$ beam search results. Average length of dialogues is employed as an evaluation metric as in \cite{li2016deep}.
		
		Machine-machine simulation is conducted in a way similar to \citep{li2016deep} in which we let two bots equipped with the same model talk with each other in $1000$ simulated dialogues. Each dialogue is initialized with the first utterance of a test example, and terminated according to the termination conditions for reward estimation in Section \ref{RL}. In human-machine conversation, we recruit $5$ native speakers as testers and ask them to talk with the bots equipped with the three models. Every time, a bot is randomly picked for a tester, and the tester does not know which model is behind. Every tester finishes $100$ dialogues with each bot. To make a fair comparison, we let the bots start dialgoues. A starting message in a dialogue is randomly sampled from the test data and copied $3$ times for all the $3$ bots. %(a tester can skip the message if he/she cannot understand it).  
		A dialogue is terminated if (1) the tester thinks the conversation cannot be continued (e.g., due to bad relevance or repetitive content etc.); or (2) the bot gives repetitive responses in two consecutive turns (measured by $cosine(e(u_{i-1}),e(u_{i+1})) > 0.9$). %Dialogue acts in human turns are tagged by the classifier in Section \ref{DiaActCla}. 
		The evaluation metric is calculated with the total $500$ dialogues for each model. 
		
		Table \ref{evalres} (b) reports the evaluation results. In both experiments, SL-DAGM and RL-DAGM can lead to longer conversations, and the improvements from both models over the baseline are statistically significant (t-test, p-value $<0.01$). Improvements in human-machine conversation are smaller than those in machine-machine simulation,  indicating the gap between the simulation environment and the real conversation environment and encouraging us to consider online optimization in human-machine conversations in the future. RL-DAGM is better than SL-DAGM in both experiments, indicating the efficacy of reinforcement learning.  In addition to the overall average length, we also show the distributions of average length of dialogues across different testers in human-machine conversation in Figure \ref{boxgraph}.  Although there exists variance among the testers, the overall trend is consistent with the numbers in Table \ref{evalres} (b).
		
		The reason that our models are better is that they captured conversational patterns in human-human interactions and obtained further optimization through reinforcement learning. First, the models can pro-actively switch contexts in a smooth way. In machine-machine simulation, $65.4$\% (SL) and $94.4$\% (RL) dialogues contain at least one CS.*; and in human-machine conversation, the two percentages are $38.1$\% (SL) and $48.1$\% (RL) respectively. More interestingly, in machine-machine simulation, average lengths of dialogues without CS.* are only $4.78$ (SL) and $2.67$ (RL) respectively which are comparable with or even worse than RL-S2S, while average lengths of dialogues with CS.* are $8.66$ (SL) and $8.18$ (RL) respectively. The results demonstrate the importance of context switch for engagement in open domain conversation and one signficant effect of RL is promoting context switch in interactions for future engagment even with sacrifice on relevance of the current turn (e.g., more 0 responses than SL-DAGM in Table \ref{evalres} (a)). Second,  the models can drive conversations by asking questions. In machine-machine simulation, $36.5$\% (SL) and $32.4$\% (RL) dialogues contain at least one question. The percentages in human-machine conversation are $17.7$\% (SL) and $22.3$\% (RL) respectively.  We show examples of machine-machine simulation and human-machine conversation in Appendix.

		%\subsection{Discussion} 		
%\vspace{-3mm}
\section{Related Work}%\vspace{-3mm}
A common practice for building an open domain dialogue model is to learn a generative model in an end-to-end fashion. On top of the basic sequence-to-sequence with attention architecture \citep{vinyals2015neural,shang2015genBased}, various extensions have been proposed to tackle the ``safe response'' problem \citep{li2015diversity,mou2016sequence,xing2017topic}; to model complicated structures of conversational contexts \citep{serban2015building,sordoni2015neural,xing2017hierarchical}; to bias responses to some specific persona or emotions \citep{li2016persona,zhou2017emotional}; and to pursue better optimization strategies \citep{li2017adversarial,li2016deep}. In this work, we consider open domain dialogue generation with dialogue acts. Unlike task-oriented dialogue systems \cite{young2013pomdp,wen2016network} where task specific dialogue acts have been extensively applied for dialogue management, only a little work on open domain dialogue modeling takes dialogue acts into account. Most of the existing work stops at performing utterance classification or clustering \citep{kim2010classifying,kim2012classifying,ivanovic2005dialogue,wallace2013generative,ritter2010unsupervised}.  Recently, \citet{zhao2017learning} incorporate dialogue acts in the Switchboard Corpus as prior knowledge into dialogue generation. \citet{serban2017deep} leverage dialogue acts as features in their response selection model. Our work is unique in that we design special dialogue acts to explain social interactions, control open domain response generation, and thus guide human-machine conversations.

		%\vspace{-3mm}
		\section{Conclusion}%\vspace{-3mm}
		We design dialogue acts to describe human behavior in social interactions and propose open domain dialogue generation with the dialogue acts as policies. The dialogue model is learned through a supervised learning approach a reinforcement learning approach. Empirical studies show that the proposed models can significantly outperform state-of-the-art methods in terms of both response quality and user engagement. 
\bibliography{acl2018}
\bibliographystyle{acl_natbib}

\section{Appendix}

\subsection{Generation Network}\label{GN}

Suppose that $x_i=[a_i;u_{i-1};u_{i-2}]=(w_{i,1},\ldots, w_{i,n'_i})$ where $w_{i,k}$ is the embedding of the $k$-th word, then the $k$-th hidden state of the encoder is given by $v_{i,k}=[\overrightarrow{v}_{i,k}; \overleftarrow{v}_{i,k}]$ where
\begin{align*}
& \overrightarrow{v}_{i,k}=f^e_{\text{GRU}} (\overrightarrow{v}_{i,k-1}, w_{i,k});\\ & \overleftarrow{v}_{i,k}=f^e_{\text{GRU}} (\overleftarrow{v}_{i,k+1}, w_{i,k})
\end{align*} 
Positions of $u_{-1}$ and $u_{0}$ in $x_1$ and $x_2$ are padded with zeros.  Let $r_i=(w'_{i,1},\ldots, w'_{i,T})$, then in decoding the $j$-th word $w'_{i,j}$, $\{v_{i,1},\ldots,v_{i,n'_i}\}$ is summarized as a context vector $c_{i,j}$ through an attention mechanism:
\begin{align*}
& c_{i,j} = \sum_{k=1}^{n'_i} \alpha_{j,k} v_{i,k}; \thickspace \alpha_{j,k} = \frac{exp(e_{j,k})}{\sum_{m=1}^{n'_i} exp(e_{j,m})}; \\ 
& e_{j,k} = v^\top tanh(W_{\alpha}[v_{i,k};v'_{i,j-1}]),
\end{align*}
where $v$ and $W_{\alpha}$ are parameters, and $v'_{i,j-1}$ is the $(j-1)$-th hidden state of the decoder GRU in which $v'_{i,j}$ is calculated by
\begin{equation*}
v'_{i,j}=f^d_{\text{GRU}} (v'_{i,j-1}, w'_{i,j-1}, c_{i,j}).
\end{equation*}
The generation probability of $w_{i,j}$ is then defined as 
\begin{equation*}
p_r(w'_{i,j} | w'_{i,<j}, x_i)=\mathcal{I}(w'_{i,j})^{\top} softmax(w'_{i,j-1}, v'_{i,j}),
\end{equation*}
where $\mathcal{I}(w'_{i,j})$ is a vector with only one element $1$ indicating the index of $w'_{i,j}$ in the vocabulary. $p_r(r_i | a_i, u_{i-1}, u_{i-2})$ is finally defined  as
\begin{equation*}
p_r(w'_{i,1}| x_i)\prod_{j=2}^T p_r(w'_{i,j}| w'_{i,<j}, x_i).
\end{equation*}

\subsection{Implementation Details of the Dialogue Act Classifier}\label{ClassLearn}
We randomly split the $500$ labeled dialogues as $400$, $30$, and $70$ dialogues for training, validation, and test respectively. Utterances in the three sets are $3280$, $210$, and $586$ respectively. In training, we represent dialogue acts as probability distributions by averaging the labels given by the three annotators. For example, if an utterance is labeled as ``CM.S'', ``CM.S'', and ``CS.S'', then the probability distribution is $(0.67,0,0,0.33,0,0,0)$. In test, we predict the dialogue act of an utterance $u_i$ by $\arg\max_{j} g(u_i,u_{i-1},a_{i-1})[j]$. To avoid overfitting, we pre-train word embeddings using word2vec\footnote{\url{https://code.google.com/archive/p/word2vec/}} with an embedding size of $200$ on the $30$ million data and fix them in training. We set the embedding size of the dialogue acts and the hidden state size of the biGRUs as $100$, and the dimensions of the first layer and the second layer of the MLP as $200$ and $7$ respectively. We optimize the objective function (i.e., Equation (3) in the submission) using back-propagation and the parameters are updated by stochastic gradient descent with AdaDelta algorithm \citep{zeiler2012adadelta}. The best performing model on the validation data is picked up for test.

\subsection{Implementation Details of the Dialogue Model}\label{expdetail}
In learning of the generation network, we set the size of word embedding  as $620$ and the size of hidden vectors as $1024$ in both the encoder and the decoder. Both the encoder vocabulary and the decoder vocabulary contain $30,000$ words. Words out of the vocabularies are replaced by a special token ``UNK".  We employ AdaDelta algorithm \citep{zeiler2012adadelta} to train the generation network with a batch size $128$. We set the initial learning rate as $1.0$ and reduce it by half if  perplexity on validation begins to increase. We stop training if the perplexity on validation keeps increasing in two successive epochs.

In learning of the policy network, we set the size of word embedding, the size of dialogue act, and the size of hidden states of the biGRU as $100$. There are $50$ neurons in the first layer of the MLP and $7$ neurons in the second layer of the MLP.  Vectors in the policy network have smaller sizes than those in the generation network because the complexity of dialogue act prediction is much lower than language generation. 

%In learning of the policy network, we set the size of word embedding and the hidden state of the dual LSTM as $100$. We randomly sample a negative instance for each instance in the 10 million training data set to form the entire training corpus for the matching model. The validation data set is processed in the same way. We use Adam \cite{kingma2014adam} algorithm to optimize the model and stop training if the performance on the validation data set stop increasing in two consecutive epochs. 

In reinforcement learning, the size of mini-batch is $60$ and learning rate is fixed as $0.05$. To estimate the reward, we train a dual LSTM \citep{lowe2015ubuntu} with the size of word embedding and the size of hidden states as $100$. Responses from the simulated dialogues are generated with a beam size $20$. 

In RL-S2S, we define $8$ responses as dull responses according to the frequency of responses in the training set. Table \ref{safeR} gives the responses.
\begin{table*}[h]
	\small
	\centering
	
	\begin{tabular}{c|l|l}
		\hline
		No.& Chinese responses & English translations\\
		\hline
		1& 我不知道  & I do not know. \\
		2& 我觉得你说得对 & I think you are right.\\
		3& 你是男的女的 & Are you a man or a woman?	\\ 
		4& 嗯 我知道 & I see.	\\ 
		5& 我也不知道 &  I do not know either.	\\ 
		6& 你说的对& You are right.	\\ 
		7& 我也是这么想的 & I think so.	\\ 
		8& 好啊& OK.	\\  		
		\hline
		
		\hline
	\end{tabular}
	\caption{Dull responses for learning RL-S2S. 
		\label{safeR} 
	}		 	
	
\end{table*}

\subsection{More Examples of Response Generation}\label{exp4RG}
We compare SL-DAGM and RL-DAGM with baseline models in terms of response quality for given contexts with more examples in Table \ref{case4}.

\begin{table*}
	\small
	\centering		
	\renewcommand\arraystretch{1.2}	
	\begin{tabular}{p{0.35\linewidth}|m{0.6\linewidth}}
		\hline
		Context &  Responses \\	\hline
		
		\begin{tabular}{b{4.5cm}}
			你眼袋略大，常熬夜?  $\Rightarrow$ 我大学毕业后没有熬夜过？ $\Rightarrow$  早睡早起好孩子 $\Rightarrow$  是的 \\
			\\
			Your have dark circles under your eyes. Always stay up late?   $\Rightarrow$  I never stay up late after I graduate from my university.  $\Rightarrow$  Good boy! Early to bed, early to rise. $\Rightarrow$  Yes.   
		\end{tabular}
		&
		\begin{tabular}{m{8cm}}
			\textbf{S2SA}: 我也是这么想的。 I think so \\
			\textbf{HRED}： 哈哈。 Haha. \\
			\textbf{VHRED}： 你是在夸我吗？ So you are praising me? \\
			\textbf{RL-S2S}： 我也是好孩子。 I am a good boy too. \\
			\textbf{SL-DAGM} ：我经常倒时差睡眠不好。 But I often have poor sleep because of jet lag. (\textbf{CS.S})\\ 
			\textbf{RL-DAGM} ：想问你每天几点起床。 Then when do you get up everyday? (\textbf{CS.Q})\\   
			\textbf{CM.Q}： 你要睡了吗？ Are you going to sleep? \\
			\textbf{CM.S}： 我也是早睡。 I sleep early as well.\\ 
		\end{tabular}\\
		\hline
		
		\begin{tabular}{b{4.5cm}}
			我们一起合租怎么样?  $\Rightarrow$ 多少钱？ $\Rightarrow$  800 $\Rightarrow$  加我微信 \\
			\\
			How about renting an apartment together?   $\Rightarrow$  How much? $\Rightarrow$800. $\Rightarrow$  Then add me as a contact in WeChat.
		\end{tabular}
		&
		\begin{tabular}{m{8cm}}
			\textbf{S2SA}: 邮件发你了。 I have sent you an email. \\
			\textbf{HRED}： 你加我微信可以吗？ Could you please add me in your WeChat? \\
			\textbf{VHRED}： 加我微信。 Add me in your WeChat. \\
			\textbf{RL-S2S}： 邮件发你了。 I have sent you an email. \\
			\textbf{SL-DAGM}： 我想和你一起玩桌游。 I want to play board games with you. (\textbf{CS.S})\\ 
			\textbf{RL-DAGM}： 什么时候联系。 When can I contact you? (\textbf{CM.Q})\\   
			\textbf{CS.Q}： 你在哪个区？ Which district do you live in? \\
			\textbf{CM.S}： 给我微信号。 Give me your WeChat ID.\\ 
		\end{tabular}\\
		\hline
	\end{tabular}	
	\caption{More examples of response generation. Utterances in the context are split by ``$\Rightarrow$''.
	}		\label{case4} 	
\end{table*}

\subsection{Examples in Engagment Test} \label{simul}

Table \ref{case2} gives some examples on machine-machine simulation. Unlike the dialogues from RL-S2S which quickly converge to loops, dialogues from our models smoothly move forward under the management of the dialogue acts.  The dialogue acts let us know why such responses are generated and make the simulated dialogues closer to human dialogues with moderate context continuation and jumping out of the contexts at proper timing.  Table \ref{case3} and Table \ref{case5} show some examples from the test of human-machine conversation. We denote a machine turn as ``M'' and a human turn as ``H''. After each example, we give the reason of termination in which ``EOD-H'' means the dialogue is terminated by the tester and ``EOD-R'' means the dialogue is terminated by the repetition check with the next generated turn attached. Compared to dialogues with the baseline, dialogues with our models can go deeper with much richer content, although a side-effect is that sometimes responses from CS.* might be nonsense (e.g., the first example of SL-DAGM). This sheds light on our future direction to further improve the generation network with knowledge.

\begin{table*}[t]
	\small
	\centering
	
	\begin{tabular}{m{4.3cm}|m{4.3cm}|m{4.3cm}}
		\hline
		RL-S2S &  SL-DAGM & RL-DAGM \\	\hline
		
		\begin{tabular}{m{4.2cm}}
			A: 无锡妹子你好！\\
			B: 我是无锡的，你在哪儿？ \\
			A: 我在上海。\\
			B: 我也在上海。 \\
			A: 我也在上海。 \\
			\\
			A: Hello, Wuxi girl!  \\
			B: I come from Wuxi. How about you?  \\
			A: I come from Shanghai.  \\
			B: I come from Shanghai too. \\
			A: I come from Shanghai too.
		\end{tabular}
		&
		\begin{tabular}{m{4.2cm}}
			A: 无锡妹子你好！ (CM.S)\\
			B: 你也无锡的？ (CM.Q)\\
			A: 嗯嗯，无锡的。 (CM.A)\\
			B: 我在无锡上学呢。 (CM.S)\\
			A: 无锡有什么好吃的吗？(CS.Q)\\
			\\
			A: Hello, Wuxi girl! (CM.S) \\
			B: You also come from Wuxi? (CM.Q) \\
			A: Yes, I am. (CM.A) \\
			B: I am a college student in Wuxi. (CM.S) \\
			A: Is there any delicious food in Wuxi? (CS.Q)\\
		\end{tabular}
		&
		\begin{tabular}{m{4.2cm}}
			A: 无锡妹子你好！ (CM.S)\\
			B: 你好，你也在江苏？ (CS.Q)\\
			A: 不是土著，在江苏读书。(CS.S)\\
			B: 江苏本地人很多啊。 (CM.S)\\
			A: 但是我身边都是外地人。(CM.S)\\
			\\
			A: Hello, Wuxi girl! (CM.S)  \\
			B: Hello, you also in Jiangsu province? (Wuxi is a city in Jiangsu) (CS.Q) \\
			A: I am not born in Jiangsu, but I go to college in Jiangsu. (CS.S) \\
			B: Many people in Jiangsu are natives.  (CM.S) \\
			A: But most people around me are not natives. (CM.S)\\
		\end{tabular}
		\\ \hline

		\begin{tabular}{m{4.2cm}}
			A: 去新加坡马来西亚可以学英语.\\
			B: 新加坡有好玩的吗？ \\
			A: 有好玩的吗？ \\
			B: 新加坡有什么好玩的？我想去看看。 \\
			\\
			A: I can learn English in Singapore and Malaysia. \\
			B: Are there any interesting places in Singapore? \\
			A: Are there any interesting places? \\
			B: Are there any interesting places in Singapore? I want to have a tour in Singapore.\\
		\end{tabular}
		&
		\begin{tabular}{m{4.2cm}}
			A: 去新加坡马来西亚可以学英语。 (CM.S)\\
			B: 你英文名是什么？ (CS.Q) \\
			A: 我应该可以写出来。 (CM.S)\\
			B: 那你姓什么。 (CM.S) \\
			A: 我什么时候写出来？ (CS.Q)\\
			\\
			A: I can learn English in Singapore and Malaysia. (CM.S) \\
			B: What is your English name? (CS.Q) \\
			A: I think I can write it out. (CM.S)  \\
			B: What is your family name? (CM.S) \\
			A: When shall I write my English name? (CS.Q)\\
		\end{tabular}
		
		&
		\begin{tabular}{m{4.2cm}}
			A: 去新加坡马来西亚可以学英语。 (CM.S)\\
			B: 你也想去新加坡学英语？ (CM.Q) \\
			A: 一样一样。 (CM.A)\\
			B: 你是什么专业的？ (CS.Q)\\
			A: 我们学校有什么专业？ (CM.Q)\\
			\\
			A: I can learn English in Singapore and Malaysia. (CM.S) \\
			B: You also want to go there to learn English? (CM.Q) \\
			A: Yep. (CM.A) \\
			B: What are you major in? (CS.Q) \\
			A: What majors do we have? (CM.Q)\\
		\end{tabular}
		\\\hline
	\end{tabular}		
	\caption{Comparison of simulated dialogues from different models.
	}		\label{case2} 	
\end{table*}

\begin{table*}[h]
	\small
	\centering

	\begin{tabular}{m{4.3cm}|m{4.3cm}|m{4.3cm}}
		\hline
		RL-S2S &  SL-DAGM & RL-DAGM \\	\hline
		
		\begin{tabular}{m{4.2cm}}
			M: 喝汤就饱了。  \\
			H: 可以减肥。 \\
			M: 减肥会不会很痛苦啊？ \\
			H: 不吃不喝确实很难受。\\
			M: 我都不敢吃了。 \\
			H: 那就不吃了？ \\			
			EOD-R (我也不敢吃了。) \\
			\\
			M: I am full with only soup. \\
			H: Then you can lose weight. \\
			M: Is it tough to lose weight? \\
			H: It's indeed tough when you have to diet. \\
			M: I have to diet.\\
			H: Then diet? \\
			EOD-R (I have to diet as well.)\\
		\end{tabular}
		&
		\begin{tabular}{m{4.2cm}}
			M: 喝汤就饱了。 (CM.S)\\
			H: 可以减肥。 (CS.S)\\
			M: 吃不胖的体质伤不起啊！ (CM.S)\ \\
			H: 羡慕。 (CM.A) \\
			M: 我正在增肥。 (CS.S)\\
			H: 多吃冰淇淋。 (CM.S)\\
			M: 冰与火之歌都吃过了。 (CM.S)\\
			EOD-H\\
			\\
			M: I am full with only soup. (CM.S)  \\
			H: Then you can lose weight. (CS.S)\\
			M: I really want to get a little fatter. (CM.S)\\  
			H: I am jealous of you! (CM.A)\\
			M: I am trying to gain weight.  (CS.S)\\
			H: Then have more ice cream. (CM.S)\\
			M: I even ate Song of Ice and Fire.(CM.S)\\
			EOD-H\\
		\end{tabular}
		&
		\begin{tabular}{m{4.2cm}}
			M: 喝汤就饱了。 (CM.S)\\
			H: 可以减肥。 (CS.S) \\
			M: 你吃蛋白粉？ (CS.Q) \\
			H: 我不吃蛋白粉。 (CM.A)\\
			M: 酵素粉是什么？(CS.Q) \\
			H: 减肥神物。 (CM.A) \\
			M: 我去试试！ (CM.S) \\
			H: 快去吧，吃酵素要坚持。 (CM.S)\\
			M: 药店有卖酵素粉的吗？多少钱啊？ (CS.Q) \\
			H: 有，不贵。 (CM.A)  \\
			EOD-R (请问药店有卖的吗？多少钱？) \\
			\\			
			M: I am full with only soup. (CM.S) \\
			H: Then you can lose weight. (CS.S) \\
			M: Do you eat albumen powder? (CS.Q) \\
			H: Nope. (CM.A) \\
			M: What is enzyme powder? (CS.Q) \\
			H: Something can make you slim. (CM.A)\\
			M: I want to try it! (CM.S)\\
			H: Go ahead, and do not give up. (CM.S)\\
			M: Can I get enzyme powder in pharmacy? How much is it? (CS.Q)\\
			H: Yes you can. It's not expensive. (CM.A)\\
			EOD-R (Can I get it in pharmacy? How much?)\\
		\end{tabular}
		\\
		\hline 
	\end{tabular}
	\caption{Example 1 of human-machine conversation. ``M'' means a machine turn, and ``H'' means a human turn. }	
	\label{case3} 	
\end{table*}

\begin{table*}[h]
	\small
	\centering

	\begin{tabular}{m{4.3cm}|m{4.3cm}|m{4.3cm}}
		\hline
		RL-S2S &  SL-DAGM & RL-DAGM \\	\hline				
		
		\begin{tabular}{m{4.2cm}}
			M: 你用淡妆比较好看。 \\
			H: 嗯 我听你的。 \\
			M: 谢谢你，一起努力啊。 \\
			H: 一起努力啊。  \\
			EOD-R （一起努力！） \\
			\\
			M: Light makeup looks good for you.  \\
			H: Well, you are right. \\
			M: Thank you.  Let's do it together!  \\
			H: Do it together! \\
			EOD-R (Do it together!) \\			
		\end{tabular}
		&
		\begin{tabular}{m{4.2cm}}
			M: 你用淡妆比较好看。 (CM.S) \\
			H: 嗯 我听你的。 (CM.A) \\
			M: 只用粉底和口红吧。 (CS.S)\\
			H: 你推荐个粉底。 (CM.S) \\
			M: 资生堂的不错。 (CM.S)\\
			H: 这个好。 (CM.S) \\
			M: 因为我用过一段时间，觉得很不错(CM.S)\\
			H: 那太好了。 (CM.S)\\
			M: 你用过几次？(CM.Q)\\
			H: 我没用过(CM.A)\\
			EOD-R(你用过几次？)\\
			\\
			M: Light makeup looks good for you. (CM.S) \\
			H: Well, you are right.  (CM.A) \\
			M: Only use foundation and lipstick. (CS.S)\\
			H: Recommend me a brand of foundation. (CM.S) \\
			M: Shiseido is good. (CM.S)\\
			H: That is great. (CM.S)\\
			M: Because I used to try it for sometime, and feel good. (CM.S)\\
			H: Sounds good.  (CM.S)\\
			M: How many times have you used it? (CM.Q) \\
			H: Never. (CM.A)\\
			EOD-R (How many times have you used it?) \\
		\end{tabular}
		&
		\begin{tabular}{m{4.2cm}}
			M: 你用淡妆比较好看。 (CM.S) \\
			H: 嗯 我听你的。 (CM.A) \\
			M: 画浓妆会不会气色很差？ (CS.Q)\\
			H: 不会的。 (CM.A)\\
			M: 那有什么副作用？ (CS.Q)\\
			H: 皮肤可能不好。 (CM.S) \\
			EOD-R (有什么副作用？)\\
			
			\\
			M: Light makeup looks good for you. (CM.S) \\
			H: Well, you are right.  (CM.A) \\
			M: Does heavy makeup make your complexion look bad? (CS.Q)\\
			H: Nope. (CM.S)\\
			M: Then is there any side-effect? (CS.Q)\\
			H: My skin will become worse. (CM.S) \\
			EOD-R (is there any side-effect?)\\
		\end{tabular}
		\\
		\hline
	\end{tabular}	
	\caption{Example 2 of human-machine conversation. ``M'' means a machine turn, and ``H'' means a human turn. }	
	\label{case5} 	
\end{table*}
	\end{CJK*}

\end{document}